\documentclass{article}
\usepackage{spconf,amsmath,graphicx,arydshln}

\usepackage{enumitem}
\setlist{nosep, leftmargin=14pt}

\usepackage{mwe} 

\title{Efficient 4D fMRI ASD Classification using Spatial-Temporal-Omics-based Learning Framework}
\name{Ziqiao Weng$^{\star \dagger}$ \qquad Weidong Cai$^{\dagger}$ \qquad Bo Zhou$^{\star}$}

\address{$^{\star}$ Department of Radiology, Northwestern University, Chicago, IL, USA \\
    $^{\dagger}$ School of Computer Science, The University of Sydney, Sydney, Australia
}
\begin{document}
%
\maketitle
\begin{abstract}

\end{abstract}

Autism Spectrum Disorder (ASD) is a neurodevelopmental disorder impacting social and behavioral development. Resting-state fMRI, a non-invasive tool for capturing brain connectivity patterns, aids in early ASD diagnosis and differentiation from typical controls (TC). However, previous methods, which rely on either mean time series or full 4D data, are limited by a lack of spatial information or by high computational costs. This underscores the need for an efficient solution that preserves both spatial and temporal information. In this paper, we propose a novel, simple, and efficient spatial-temporal-omics learning framework designed to efficiently extract spatio-temporal features from fMRI for ASD classification. Our approach addresses these limitations by utilizing 3D time-domain derivatives as the spatial-temporal inter-voxel omics, which preserve full spatial resolution while capturing diverse statistical characteristics of the time series at each voxel. Meanwhile, functional connectivity features serve as the spatial-temporal inter-regional omics, capturing correlations across brain regions. Extensive experiments and ablation studies on the ABIDE dataset demonstrate that our framework significantly outperforms previous methods while maintaining computational efficiency. We believe our research offers valuable insights that will inform and advance future ASD studies, particularly in the realm of spatial-temporal-omics-based learning.


\vspace{-5px}
\section{Introduction}
\label{sec:intro}

Autism Spectrum Disorder (ASD) is a neurodevelopmental disorder that impacts social interaction, communication, learning, and behavior \cite{sharma2018autism}. Early diagnosis is crucial for effective intervention and treatment but presents significant challenges. Traditional clinical diagnostic methods primarily rely on behavioral and cognitive assessments, which are not only time-consuming and labor-intensive but also highly subjective, increasing the risk of misdiagnosis. Therefore, developing fully automated ASD diagnostic technology is essential, as it would alleviate the burden on clinicians and facilitate early symptom identification, enabling timely intervention and treatment \cite{jiang2022cnng}. Functional Magnetic Resonance Imaging (fMRI) is widely valued for its non-invasiveness, high spatial-temporal resolution, and ability to provide detailed insights into both physiological and pathological brain activity \cite{park2023residual,jiang2022cnng,liu2021autism,jiang2022cnng,ziqiao}. Resting-state fMRI (rs-fMRI), which captures brain activity in a resting state, is particularly suitable for ASD patients and has been widely used to investigate altered brain connectivity and activity patterns in ASD. In line with most ASD classification studies, we leverage rs-fMRI data to distinguish ASD patients from typical controls.

Deep learning (DL) has been extensively applied to the diagnosis of brain disorders, including ASD. Given that rs-fMRI data capture both spatial and temporal signals, many DL-based approaches focus on dimensionality reduction to manage the complex spatio-temporal information effectively. These methods typically employ different brain atlases to partition the brain into approximately 100 to 400 distinct functional regions. For each region of interest (ROI), the mean time series is extracted and either directly processed by a 1D convolutional network \cite{el2019simple} or used to compute correlation coefficients between ROIs, thereby constructing a functional connectivity matrix for subsequent classification \cite{li2020multi,park2023residual,niu2020multichannel,yang2022classification,heinsfeld2018identification,li2020pooling,sherkatghanad2020automated,eslami2019asd}. Since these hand-crafted features can be precomputed and the classifiers are typically simple, such methods are computationally efficient and less prone to overfitting. However, this aggressive downscaling often fails to fully capture the rich spatio-temporal information inherent in 4D fMRI data, potentially disrupting essential temporal and spatial correlations. Consequently, the ability to detect complex neural activity patterns may be compromised, leading to suboptimal performance. Alternatively, to avoid excessive dimensionality reduction and better preserve spatio-temporal information, several studies have proposed learning directly from full 4D fMRI sequences \cite{el2019hybrid,asadi2023transformer,bengs20204d,jiang2022cnng,liu2023asd}. While this approach retains more information, processing full 4D fMRI data typically requires a significantly larger number of network parameters, resulting in higher computational costs and longer convergence times. This reduces efficiency and increases the risk of overfitting. Additionally, the varying temporal lengths of 4D data, due to different scan durations, necessitate cropping fixed-length sub-sequences during training. During inference, a temporal sliding window is used to cover the entire time series, further exacerbating inefficiencies.

In this study, we propose a spatial-temporal-omics-based learning framework (STO) to overcome the limitations of previous methods and enable efficient and intelligent learning of spatio-temporal features for ASD classification. Our framework integrates two complementary types of spatial-temporal omics: inter-voxel and inter-regional. For spatial-temporal inter-voxel omics (STVOmics), which focuses on voxel-level analysis, we extract 3D time-domain derivatives from 4D fMRI, capturing diverse statistical characteristics of the time series at each voxel while preserving full spatial resolution \cite{thomas2020classifying}. These features are then processed by a 3D CNN to generate rich voxel-wise representations. In contrast, spatial-temporal inter-regional omics (STROmics), which focuses on regional-level analysis, utilizes functional connectivity features to produce compact region-wise information through a single-layer perceptron (SLP). The voxel-wise and region-wise omics are subsequently fused and passed through an SLP for final ASD classification. Furthermore, our method is highly scalable, allowing seamless integration with existing functional connectivity-based techniques for inter-regional omics analysis.

We evaluated our approach against methods that either process full 4D fMRI or rely on mean time series across varying dataset sizes using the ABIDE benchmark dataset for ASD classification. Our results demonstrate that the proposed framework consistently achieves significant performance improvements over other methods in all scenarios while maintaining simplicity and computational efficiency. We believe our experiments and analyses provide valuable insights for advancing ASD research.

\vspace{-10px}
\section{Method}
\label{sec:method}

\subsection{Spatial-Temporal-Omics-based Framework}
Previous DL-based methods for ASD classification either extract mean time series using various brain atlases or explore spatio-temporal information from full 4D fMRI data. The first approach often struggles to capture complex neural activity patterns due to significant spatial information loss, while the second approach increases computational cost and model complexity. To address these issues, our STO framework provides an effective solution by combining two complementary omics. This method preserves spatial resolution while extracting rich temporal features. As shown in Figure 1, the framework integrates inter-voxel and inter-regional omics, processed in parallel branches.

\noindent\textbf{Spatial-Temporal Inter-Voxel Omics}: We extract 3D time-domain statistical features from 4D fMRI data at each voxel, known as STVOmics. This approach preserves full spatial resolution, effectively capturing the detailed spatio-temporal characteristics of fMRI data across voxels. The extracted features are then processed by a lightweight 3D CNN to generate rich voxel-wise representations.

\noindent\textbf{Spatial-Temporal Inter-Regional Omics}: Following Li et al. \cite{li2020multi}, we select features from the upper triangle of the functional connectivity matrix as STROmics, which capture spatio-temporal characteristics across brain regions and reduce the complexity of 4D data. These 1D features are passed through a fully connected layer to be mapped into the same embedding space as the inter-voxel omics features. To demonstrate the versatility of inter-regional omics in integrating with functional connectivity-based methods, we adopt the DiagNet training paradigm \cite{eslami2019asd}. In this approach, input features are downsampled, and a decoder is introduced after the encoder to reconstruct the input from intermediate features, as shown in the reconstruction section of Figure 1.

Finally, the encoded features from both omics are concatenated and fed into a fully connected layer to generate predictions for ASD classification.

\begin{figure}[htb]
\begin{minipage}[b]{1.0\linewidth}
    \centering
    \centerline{\includegraphics[width=8.5cm]{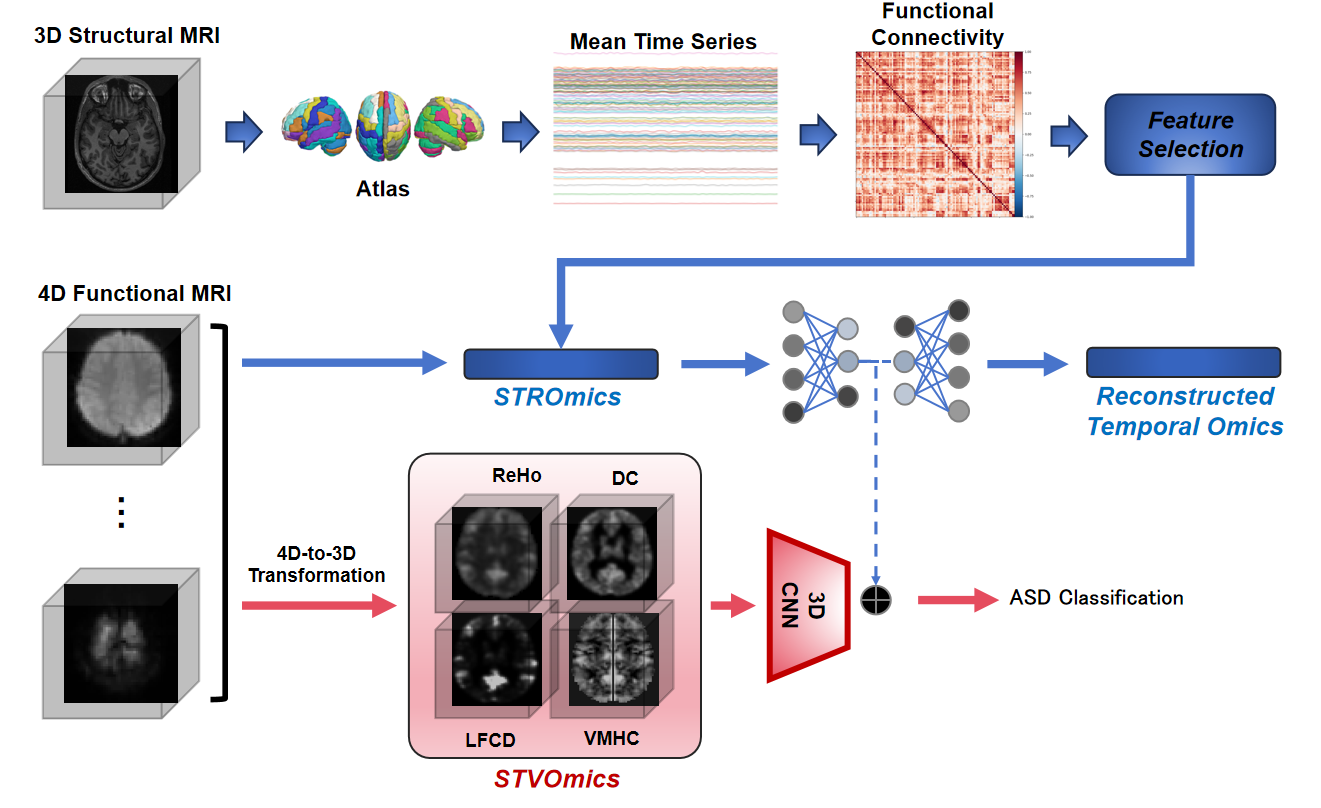}}
    
 \caption{\small{Overview of our STO framework, consisting of (a) spatial-temporal inter-regional omics
 (STROmics) and (b) spatial-temporal inter-voxel omics (STVOmics). }}
	\label{fig:framework}
	\vspace{-15px}
 \end{minipage}
 \vspace{-10px}
\end{figure}

\vspace{-5px}
\subsection{Dataset and Preprocessing}

This study leveraged the publicly available ABIDE-I dataset provided \cite{di2014autism}, which comprises data from 1,112 subjects, including 539 ASD patients and 573 healthy controls, collected from 17 different sites across North America and Europe. The dataset includes T1-weighted structural brain images, fMRI scans, and extensive phenotypic information for each subject. After performing rigorous quality control, we finalized a dataset of 871 subjects, consisting of 403 ASD patients and 468 healthy controls. The dataset was then preprocessed using the Connectome Computation System (CCS) pipeline, which involved several key steps: slice timing correction, motion correction, and voxel intensity normalization. Subsequently, nuisance signal removal was performed to mitigate confounding variations caused by head motion, physiological processes. After nuisance regression, bandpass filtering (0.01–10 Hz) was applied, without global signal correction. Finally, spatial normalization was conducted to align the brain images with the Montreal Neurological Institute (MNI) template, achieving a uniform resolution of \(3 \times 3 \times 3\) mm³. It is noteworthy that while the 3D spatial dimensions of the fMRI data are consistent across sites \(61 \times 73 \times 61\), the temporal dimensions vary from 78 (OHSU) to 316 (CMU).

In addition to the 4D fMRI data, the ABIDE initiative offers several valuable statistical derivatives and mean time series for different brain atlases. For methods based on functional connectivity, we utilized the widely adopted Automated Anatomical Labeling (AAL) atlases and the Craddock 200 (CC200) atlas to compute the average time series across 116 and 200 functionally homogeneous ROIs. The resulting mean time series is organized into a 2D matrix, with each row representing a time point and each column corresponding to an ROI. For temporal statistics, we employed four 3D derivatives that preserve the full spatial resolution of the original 4D fMRI data. These derivatives include Regional Homogeneity (ReHo), Degree Centrality (DC), Local Functional Connectivity Density (LFCD), and Voxel-Mirrored Homotopic Connectivity (VMHC). Further details regarding these derivatives can be found in \cite{craddock2013towards}.

\vspace{-10px}
\subsection{Evaluation Metrics and Baseline}
Following most studies on ASD classification, we performed 5-fold cross-validation. We applied data splitting separately to the ASD and TC subsets in each fold to ensure that class proportions remained consistent with the original dataset in both the training and testing folds.


We adopt the Area Under the ROC Curve (AUC) metric for evaluating classification performance. Unlike accuracy and F1 score, which are highly sensitive to decision boundary selection, AUC offers a more robust and interpretable assessment across varying thresholds. In addition, we report model parameters, GFLOPs, and memory consumption to comprehensively evaluate computational efficiency.


To ensure a fair comparison, we replicated several representative methods based on mean time series and 4D data, using the same data and settings. Additionally, to demonstrate the robustness and effectiveness of our proposed framework, we evaluated all methods with varying data proportions (100\%, 75\%, 50\%), modifying only the number of training samples while keeping the same test set across experiments. Below, we provide a brief overview of the baseline methods implemented for comparison with our proposed method:

\noindent\textbf{Baselines with Mean Time Series as Input}:
\begin{itemize}
    \item 1DConv \cite{el2019simple}: Utilizes a 1D convolutional layer to process the mean time series matrix, where each channel corresponds to a different ROI, and the signal sequence captures the entire temporal progression.
    
    \item FC\_2DCNN \cite{sherkatghanad2020automated}: Takes the functional connectivity matrix as input to a 2D CNN. This network consists of 7 parallel blocks, each containing a convolutional layer and a max-pooling layer. The convolutional kernel sizes range from (1, M) to (7, M), where M is the number of ROIs, operating on rows that represent brain regions.
    
    \item FC\_MLP \cite{li2020multi}: Retains only the upper triangle of the correlation matrix, flattens these values into 1D vectors, and uses them as input to a two-layer MLP. 
    
    \item DiagNet \cite{eslami2019asd}: Downsamples the upper triangle of the correlation matrix by selecting the top and bottom 1/4 of correlations, discarding the rest. An autoencoder is then trained to minimize reconstruction error and further reduce feature size. The encoded features are passed to a SLP for final ASD classification. The autoencoder and classifier are trained simultaneously.
\end{itemize}

\noindent\textbf{Baselines with Full 4D Data as Input}:
For these methods, we set the cropped time steps during training and the sliding stride during inference to 15, while reducing the spatial resolution to 32, consistent with the 3D approaches. 
\begin{itemize}
    \item 3DCNN-MS \cite{li20182}: Utilizes a sliding window to compute temporal statistics (mean and standard deviation) as 2-channel inputs to a 3D CNN (ResNet-10).
    \item 3DCNN-TC \cite{bengs20204d}: Treats time points as stacked channels and processes the 4D data using a 3D CNN (ResNet-10). 
    \item ConvGRU-CNN3D \cite{bengs20204d}: First applies temporal processing with a 3D ConvGRU, followed by spatial processing with a 3D CNN (ResNet-10).
    \item CNN3D-GRU \cite{jiang2022cnng}: Combines a 3D CNN with a GRU for spatio-temporal feature extraction.
    \item 3DCNN-ConvLSTM \cite{el2019hybrid}: Uses a shallow 3D CNN with four convolutional layers at each time point, followed by a two-layer bidirectional ConvLSTM to capture both local spatial and global temporal information. The resulting 3D feature maps are processed by a 3D CNN, temporally pooled, and then passed through a fully connected layer for classification.

\end{itemize}

\vspace{-10px}
\subsection{Implementation}
Without loss of generality, we chose the 3D ResNet-10 to process the STVOmics. The network starts with an initial convolutional layer, omitting max-pooling and using a kernel size of 3. It is followed by three Basic-Res-Blocks, each containing two \(3 \times 3 \times 3\) convolutional layers, and ends with a \(1 \times 1 \times 1\) downsampling layer. Global average pooling is applied after the final module to obtain spatial features. The encoded feature dimensionality for both omics is standardized to 512. The final fully connected layer uses a dropout rate of 0.2, followed by a sigmoid activation for ASD classification probabilities.

To improve the efficiency and reduce computational cost, we downscaled the spatial dimensions of the 3D temporal statistics to \(32 \times 32 \times 32\), aligning with our implementation of full 4D fMRI-based methods. To enhance generalization, we applied standard spatial augmentations—flipping, rotation, translation, and scaling—to the 3D statistics. All models were trained to convergence with a batch size of 8. We used the Adam optimizer with a learning rate of $1e{-5}$ to minimize cross-entropy loss. During training, performance was evaluated on the validation set every 5 epochs, and the best-performing model was selected for final evaluation.

\vspace{-5px}
\section{Experiments and Results}
\label{sec:experiments}
\vspace{-5px}
We compared our models to previous deep learning-based ASD classification approaches, each employing distinct strategies for handling temporal and spatial information.

\vspace{-15px}
\begin{table}[h]
\centering
\caption{\small{The comparison results of three methods across different data proportions. \textbf{Best results}; \underline{second-best results}.}}
\renewcommand{\arraystretch}{1.4}
\resizebox{\columnwidth}{!}{%
\begin{tabular}{|lcccccc|}
\hline
\multicolumn{1}{|l|}{Method}            & \multicolumn{1}{c|}{100\%}                & \multicolumn{1}{c|}{75\%}                 & \multicolumn{1}{c|}{50\%}        
& \multicolumn{1}{c|}{Param (M)} & \multicolumn{1}{c|}{GFLOPs} & \multicolumn{1}{c|}{Memory (MB)}

\\ \hline
\multicolumn{7}{|c|}{4D fMRI Based}                                                                                                                                                               \\ \hline
\multicolumn{1}{|l|}{CNN3D-MS}          & \multicolumn{1}{c|}{0.634/0.040}          & \multicolumn{1}{c|}{0.644/0.022}          & \multicolumn{1}{c|}{0.614/0.047}     
& \multicolumn{1}{c|}{14.337} & \multicolumn{1}{c|}{12.2178} & \multicolumn{1}{c|}{56.48}
\\ \hdashline
\multicolumn{1}{|l|}{CNN3D-TC}          & \multicolumn{1}{c|}{0.631/0.020}          & \multicolumn{1}{c|}{0.628/0.021}          & \multicolumn{1}{c|}{0.599/0.052}    
& \multicolumn{1}{c|}{14.359} & \multicolumn{1}{c|}{12.9539} & \multicolumn{1}{c|}{58.19}
\\ \hdashline
\multicolumn{1}{|l|}{ConvGRU-CNN3D}     & \multicolumn{1}{c|}{0.655/0.040}          & \multicolumn{1}{c|}{0.655/0.023}          & \multicolumn{1}{c|}{0.632/0.038}  
& \multicolumn{1}{c|}{14.488} & \multicolumn{1}{c|}{84.3216} & \multicolumn{1}{c|}{58.68}
\\ \hdashline
\multicolumn{1}{|l|}{CNN3D-GRU}         & \multicolumn{1}{c|}{0.653/0.024}          & \multicolumn{1}{c|}{0.632/0.038}          & \multicolumn{1}{c|}{0.604/0.034}     
& \multicolumn{1}{c|}{0.189} & \multicolumn{1}{c|}{2.2419} & \multicolumn{1}{c|}{15.24}
\\ \hdashline
\multicolumn{1}{|l|}{CNN3D\_C-LSTM}     & \multicolumn{1}{c|}{0.670/0.041} & \multicolumn{1}{c|}{0.670/0.043}          & \multicolumn{1}{c|}{0.637/0.042}                    
& \multicolumn{1}{c|}{239.543} & \multicolumn{1}{c|}{30.4100} & \multicolumn{1}{c|}{923.80}
\\ \hline

\multicolumn{7}{|c|}{Mean Time Series Based (AAL)}                                                                                                                          \\ \hline

\multicolumn{1}{|l|}{1DConv}            & \multicolumn{1}{c|}{0.609/0.036}          & \multicolumn{1}{c|}{0.569/0.039}          & \multicolumn{1}{c|}{0.551/0.006}    
& \multicolumn{1}{c|}{0.041} & \multicolumn{1}{c|}{0.0082} & \multicolumn{1}{c|}{8.37}
\\ \hdashline
\multicolumn{1}{|l|}{FC-CNN2D}            & \multicolumn{1}{c|}{0.675/0.040}          & \multicolumn{1}{c|}{0.667/0.033}          & \multicolumn{1}{c|}{0.648/0.036}       
& \multicolumn{1}{c|}{0.378} & \multicolumn{1}{c|}{0.0422} & \multicolumn{1}{c|}{9.62}
\\ \hdashline
\multicolumn{1}{|l|}{FC-MLP}          & \multicolumn{1}{c|}{0.694/0.029}          & \multicolumn{1}{c|}{0.670/0.029}          & \multicolumn{1}{c|}{0.675/0.023}          
& \multicolumn{1}{c|}{0.107} & \multicolumn{1}{c|}{0.0001} & \multicolumn{1}{c|}{8.56}
\\ \hdashline
\multicolumn{1}{|l|}{DiagNet}           & \multicolumn{1}{c|}{0.706/0.036} & \multicolumn{1}{c|}{0.675/0.032} & \multicolumn{1}{c|}{\underline{0.679/0.031}}           
& \multicolumn{1}{c|}{11.122} & \multicolumn{1}{c|}{0.0111} & \multicolumn{1}{c|}{52.18}
\\ \hline

\multicolumn{1}{|l|}{STO}           & \multicolumn{1}{c|}{\textbf{0.739/0.045}} & \multicolumn{1}{c|}{\textbf{0.712/0.030}} & \multicolumn{1}{c|}{\textbf{0.709/0.025}}    
& \multicolumn{1}{c|}{17.757} & \multicolumn{1}{c|}{12.3345} & \multicolumn{1}{c|}{78.96}

\\ \hdashline

\multicolumn{1}{|l|}{STO (DiagNet)}           & \multicolumn{1}{c|}{\underline{0.715/0.029}} & \multicolumn{1}{c|}{\underline{0.691/0.019}} & \multicolumn{1}{c|}{\underline{0.694/0.036}}     
& \multicolumn{1}{c|}{17.759} & \multicolumn{1}{c|}{12.3345} & \multicolumn{1}{c|}{77.93}

\\ \hline

\multicolumn{7}{|c|}{Mean Time Series Based (CC200)}                                                                                                                          \\ \hline

\multicolumn{1}{|l|}{1DConv}            & \multicolumn{1}{c|}{0.608/0.043}          & \multicolumn{1}{c|}{0.590/0.042}          & \multicolumn{1}{c|}{0.541/0.019}                    
& \multicolumn{1}{c|}{0.121} & \multicolumn{1}{c|}{0.0242} & \multicolumn{1}{c|}{8.74}
\\ \hdashline

\multicolumn{1}{|l|}{FC-CNN2D}            & \multicolumn{1}{c|}{0.709/0.036}          & \multicolumn{1}{c|}{0.677/0.025}          & \multicolumn{1}{c|}{0.690/0.030}                    
& \multicolumn{1}{c|}{1.123} & \multicolumn{1}{c|}{0.2195} & \multicolumn{1}{c|}{12.56}
\\ \hdashline

\multicolumn{1}{|l|}{FC-MLP}          & \multicolumn{1}{c|}{0.721/0.040}          & \multicolumn{1}{c|}{0.711/0.034}          & \multicolumn{1}{c|}{0.712/0.027}           
& \multicolumn{1}{c|}{0.318} & \multicolumn{1}{c|}{0.0003} & \multicolumn{1}{c|}{9.42}
\\ \hdashline

\multicolumn{1}{|l|}{DiagNet}           & \multicolumn{1}{c|}{0.728/0.035} & \multicolumn{1}{c|}{0.710/0.033} & \multicolumn{1}{c|}{0.714/0.022}           
& \multicolumn{1}{c|}{99.022} & \multicolumn{1}{c|}{0.0990} & \multicolumn{1}{c|}{385.94}
\\ \hline

\multicolumn{1}{|l|}{STO}           & \multicolumn{1}{c|}{\underline{0.744/0.021}} & \multicolumn{1}{c|}{\underline{0.723/0.020}} & \multicolumn{1}{c|}{\underline{0.725/0.025}}         & \multicolumn{1}{c|}{24.531} & \multicolumn{1}{c|}{12.3413} & \multicolumn{1}{c|}{103.88} 
\\ \hdashline

\multicolumn{1}{|l|}{STO (DiagNet)}           & \multicolumn{1}{c|}{\textbf{0.752/0.020}} & \multicolumn{1}{c|}{\textbf{0.730/0.028}} & \multicolumn{1}{c|}{\textbf{0.725/0.022}}         & \multicolumn{1}{c|}{24.541} & \multicolumn{1}{c|}{12.3413} & \multicolumn{1}{c|}{105.05} 
\\ \hline

\end{tabular}
}
\end{table}

\vspace{-10px}
We report the AUC (mean/std) of three methods across different data proportions (100\%, 75\%, 50\%) as shown in Table 1. By leveraging functional connectivity features from the CC200 atlas as the temporal signal, and adopting a training paradigm similar to DiagNet, our method consistently achieves the highest mean AUCs across all scenarios: 0.752 (100\%), 0.730 (75\%), and 0.725 (50\%). These results surpass the best mean time series-based method (DiagNet with the CC200 atlas) by 2.4\%, 2.0\%, and 1.1\%, and outperform the best 4D fMRI-based method (CNN3D C-LSTM) by 8.2\%, 6.0\%, and 8.8\%. Furthermore, our method consistently performs better than others regardless of the atlas used (AAL or CC200), with significant improvements seen in both the vanilla and DiagNet versions over their corresponding baselines (FC-MLP and DiagNet). For example, the vanilla version using the AAL atlas exceeds FC-MLP by 4.5\%, 4.2\%, and 3.4\% across the three dataset proportions. These findings clearly demonstrate the superiority and robustness of our STO framework, showcasing its ability to significantly enhance functional connectivity methods. 

Our method demonstrates significantly higher efficiency, with fewer network parameters, lower GFLOPs, and reduced memory usage compared to CNN3D C-LSTM, while also achieving a substantial reduction in parameters relative to DiagNet with the CC200 atlas. Additionally, both our approach and the mean time series-based method perform inference on a single data point in under 0.1 seconds, whereas CNN3D C-LSTM takes over 0.3 seconds for a single crop, not to mention the time required for the sliding window process across the entire time series. A comparison between mean time series-based methods and 4D fMRI-based methods shows that the former consistently delivers better performance with significant computational advantages. This underscores the effectiveness of brain functional connectivity and its robustness against overfitting, validating our choice of functional connectivity as the STROmics in the STO framework.



To identify the optimal STVOmics, we conducted ablation experiments using 3D temporal statistics. While the ABIDE dataset provides eight types of time-domain derivatives, for simplicity, we focused on four key derivatives: ReHo, DC, LFCD, and VMHC. We assessed the performance of each derivative individually and evaluated the combined effect by concatenating them along the channel dimension. As shown in Table 2, the results align with our expectations, demonstrating that the combination of four derivatives yields the best performance. Furthermore, a comparison of the results in Table 1 with those in Table 2 shows that our framework—combining both STVOmics and STROmics—consistently outperforms the use of either omics alone, emphasizing the complementary nature of these two types.
\vspace{-20px}
\begin{table}[h]
\centering
\caption{\small{Ablation on 3D temporal statistics.}}
\resizebox{\columnwidth}{!}{%
\begin{tabular}{|lccc|}
\hline
\multicolumn{1}{|l|}{Method}            & \multicolumn{1}{c|}{100\%}                & \multicolumn{1}{c|}{75\%}                 & \multicolumn{1}{c|}{50\%}                                  \\ \hline

\multicolumn{1}{|l|}{ReHo}              & \multicolumn{1}{c|}{0.698/0.024}          & \multicolumn{1}{c|}{0.692/0.019} & \multicolumn{1}{c|}{0.670/0.027}                   \\ \hdashline
\multicolumn{1}{|l|}{DC}                & \multicolumn{1}{c|}{0.651/0.028}          & \multicolumn{1}{c|}{0.637/0.025}          & \multicolumn{1}{c|}{0.639/0.026}                   \\ \hdashline

\multicolumn{1}{|l|}{LFCD}              & \multicolumn{1}{c|}{0.670/0.028}          & \multicolumn{1}{c|}{0.649/0.023}          & \multicolumn{1}{c|}{0.640/0.047}                   \\ \hdashline
\multicolumn{1}{|l|}{VMHC}              & \multicolumn{1}{c|}{0.667/0.024}          & \multicolumn{1}{c|}{0.656/0.013}          & \multicolumn{1}{c|}{0.638/0.036}                    \\ \hdashline

\multicolumn{1}{|l|}{ReHo-DC-LFCD-VMHC} & \multicolumn{1}{c|}{\textbf{0.717/0.026}} & \multicolumn{1}{c|}{\textbf{0.701/0.021}}          & \multicolumn{1}{c|}{\textbf{0.687/0.032}} 
\\ \hline
\end{tabular}
}
\vspace{-5px}
\end{table}

\vspace{-20px}
\section{Discussion and Conclusion}
\vspace{-10px}

We introduce a spatial-temporal-omics learning framework that addresses the limitations of previous methods relying on mean time series or 4D data. Traditional approaches often struggle with limited spatial information, increased model complexity, and high computational costs. Our framework overcomes these challenges by using STVOmics to extract 3D time-domain derivatives, preserving inter-voxel information, and STROmics to leverage 1D functional connectivity features, capturing rich inter-regional temporal dynamics with low computational cost. The synergy of these two omics significantly improves classification performance. Results on the ABIDE dataset demonstrate that our method outperforms existing approaches across all data proportions, while maintaining a simple network structure and reducing computational overhead. Although this paper presents the initial version of the STO framework, future work could incorporate additional 3D temporal statistics into STVOmics and explore more advanced methods for STROmics. A key future direction will be developing better strategies to integrate both omics and further improve performance.

\vspace{-10px}
\section{Acknowledgments}
\label{sec:acknowledgments}
\vspace{-5px}
This research was supported by Australian Government Research Training Program (RTP) scholarship.

\vspace{-5px}
\section{Compliance with Ethical Standards}
This research study was conducted retrospectively using human subject data made available in open access by \cite{di2014autism}. Ethical approval was not required as confirmed by the license attached with the open access data.

\vspace{-5px}
\small
\bibliographystyle{IEEEbib}
\bibliography{strings,refs}
\end{document}